\documentclass{article} 
\usepackage{aaai19,times}
\usepackage{hyperref}
\usepackage{url}
\usepackage{color}
\usepackage{xspace}
\usepackage{amsmath}
\usepackage{amssymb}
\usepackage{algorithm}
\usepackage{algorithmic}
\usepackage{graphicx}
\usepackage{array}
\usepackage[flushleft]{threeparttable}
\usepackage{booktabs}
\usepackage{multirow}
\usepackage{stmaryrd}
\usepackage{float}
\usepackage[caption = false]{subfig}\usepackage[numbers]{natbib}
\usepackage[nottoc]{tocbibind}
\graphicspath{ {images/} }
\usepackage{pdfcomment}

\title{Neural Machine Translation For Paraphrase Generation}

\author{
	Alex Sokolov, Denis Filimonov \\
    Amazon Echo, Seattle, WA, USA \\
    \texttt{{alexsoko,denf}@amazon.com}
}

\newcommand{\ccite}[1]{(\citeauthor{#1}, \citeyear{#1})}

\begin{document}

\maketitle

\begin{abstract}
Training a spoken language understanding system, as the one in Alexa, typically requires a large human-annotated corpus of data. Manual annotations are expensive and time consuming. In Alexa Skill Kit (ASK) user experience with the skill greatly depends on the amount of data provided by skill developer. In this work, we present an automatic natural language generation system, capable of generating both human-like interactions and annotations by the means of paraphrasing. Our approach consists of machine translation (MT) inspired encoder-decoder deep recurrent neural network. We evaluate our model on the impact it has on ASK skill, intent, named entity classification accuracy and sentence level coverage, all of which demonstrate significant improvements for unseen skills on natural language understanding (NLU) models, trained on the data augmented with paraphrases.
\end{abstract}

\section{Introduction}
\begin{figure}[h!]
     \centering
     \includegraphics[width=0.4\textwidth]{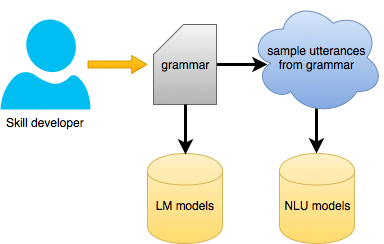}
     \caption{Skill Developer provides the grammar corresponding to the skill intents and slots. NLU models are trained on data sampled from this grammar. An example of grammar would be: \emph{''book a flight from \textless city\textgreater\ to \textless city\textgreater ''}}
     \label{fig:dev-grammar}
\end{figure}

ASK is an increasingly important part of Alexa user experience.\ccite{Kumar2017JustAB} In ASK work flow, the skill developer provides a set of \emph{slots} (often catalogs of entities), and a list of \emph{intents}, which can be mapped to actions, and a set of example phrases defining the grammar of an intent. (see figure \ref{fig:dev-grammar}).\\

\begin{figure}[h!]
     \centering
     \includegraphics[width=0.4\textwidth]{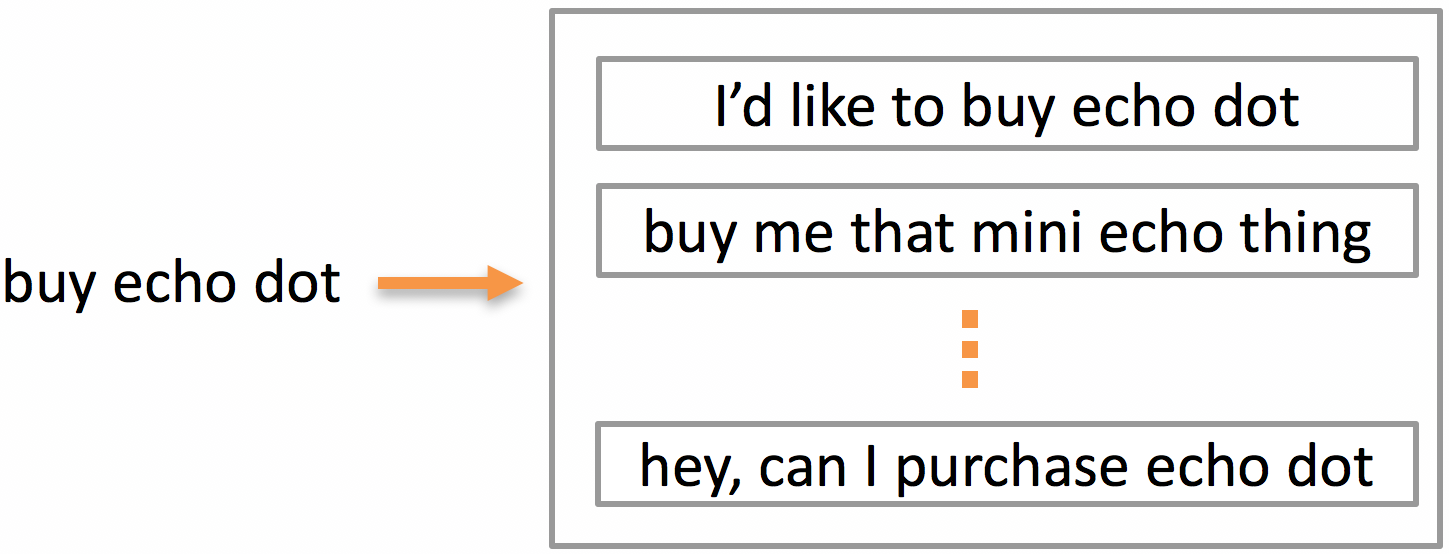}
     \caption{Desired paraphrases example. All sentences share the same entities and intent.}
     \label{fig:paraphrase}
\end{figure}

From these examples NLU and language modeling (LM) models are built for the skill. Note that it is up to the developer to anticipate all ways their users will interact with the skill. Interactions not covered by the provided examples often have much lower ASR recognition and NLU classification accuracy. Coming up with an exhaustive list of examples can be a hard task for the developer and incomplete coverage can be a frustrating experience to the user. In this work, we propose to use paraphrasing to expand the coverage of developer-provided examples, and thus reduce burden on skill developers and make skill interactions more natural to Alexa customers. Instead of relying on the developer to come up with an exhaustive list of examples for a given intent, in the proposed work flow, we will only require a few examples and then use paraphrasing model to generate other ways a customer might phrase the same command, and then use that data to build better NLU and LM models. Figure \ref{fig:paraphrase} gives an example of the desired paraphrases for a customer utterance.

Paraphrasing is used in various Natural Language Processing applications, such as natural language generation, summarization, information extraction, sentence compression and question answering. Traditional paraphrase generation methods exploit hand-crafted rules \ccite{mckeown1983paraphrasing} or automatically learned complex paraphrase patterns \ccite{zhao2009application}, use thesaurus-based \ccite{hassan2007unt} or semantic analysis driven natural language generation approaches \ccite{kozlowski2003generation}, or leverage statistical machine translation \ccite{quirk2004monolingual}; \ccite{wubben2010paraphrase}.

In this paper, we propose to use neural machine translation (NMT) as a simple and flexible approach to MT to address the paraphrase generation problem. We observe that in translation, there is not a single correct translation target, but rather several variants of the sentence, carrying the same meaning, or paraphrases. From this perspective, translation can be seen as paraphrasing the source sentence in a different language. Therefore, NMT is quite natural approach to paraphrasing. It has been shown to have comparable performance to the phrase-based translation systems \ccite{sutskever2014sequence}, and it is very flexible and modular, allowing to reuse pre-trained components, such as word embeddings or other networks trained on different datasets.

The remainder of the paper is organized as follows:  Section \ref{nmt} presents a brief overview of the sequence to sequence models and techniques used in this work, Section \ref{data} describes the available data, Section \ref{model-desc} explains the experimental setup, Section \ref{results} presents the evaluation results, Section \ref{discussion} analyzes the results and discusses future work, Section \ref{conclusion} is conclusion. 

\section{Background: Neural Machine Translation} \label{nmt}

In encoder-decoder based NMT, the meaning of the source sentence is projected into a lower dimensionality space by the encoder, from which translations (paraphrases) are generated by the decoder. It is very suitable for our needs, since it would allow to train the encoder separately on completely different dataset. This is important because we have a rather limited amount of Alexa in-domain training data, which is not parallel and has relatively small diversity and sentence length. Thus, the encoder trained on large general English dataset will produce a stronger compressed input sentence representation, which the decoder will generate output sentences from.

\begin{figure}[h!]
     \centering
     \includegraphics[width=0.55\textwidth]{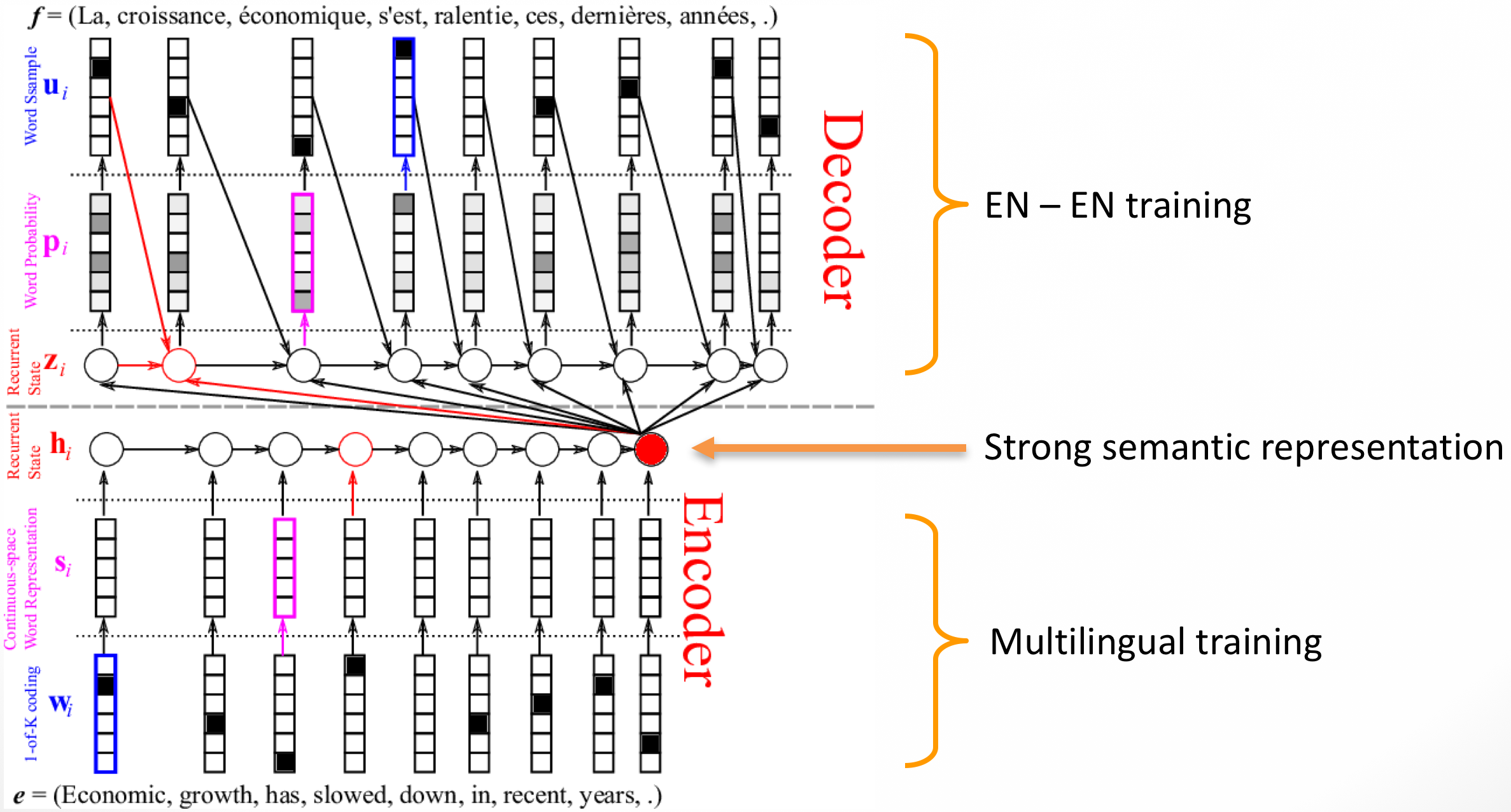}
     \caption{Encoder-decoder architecture for English to French MT model. Input and output are one-hot encoded. Horizontal axis is time, the input words are propagated through corresponding recurrent hidden states.}
     \label{fig:encdec}
\end{figure}

Encoder-decoder architecture \ccite{cho2014properties} is shown on figure \ref{fig:encdec}. Encoder operates on the source language sentence to encode it into a vector representation. The last hidden state of the encoder accumulates all the information from the sentence, which is then passed as input to the decoder. The decoder then generates output one word at a time, taking the previous generated word as input. This way the decoder gets additional signal on the past context at each generation step.

The training objective is to maximize the log probability of the target sequence given the source sequence. Therefore, the best possible decoded target is the one that has the maximum score over the length of the sequence. We perform left-to-right beam search during decoding to generate n-best paraphrase hypotheses \ccite{sutskever2014sequence}.

In addition, we use bidirectional LSTM on the first layer, so that the LSTM hidden states of each timestep summarizes not only preceding, but also the following words. It tends to perform better than regular LSTM, especially on longer sequences \ccite{schuster1997bidirectional}. We are also using GloVe \ccite{pennington2014glove} word embedding on the input. It reduces the model parameter space, since we can use a dense embedding instead of large vocabulary size one-hot encoded input. It also gives us more robustness to rare words and synonyms.

\section{Data} \label{data}
Any MT task requires parallel dataset, in our case with pairs of English input corresponding to several English outputs, such as PPDB \ccite{ganitkevitch2013ppdb}. However, there is no such dataset for Alexa domain. So we have gathered a dataset with about 400K unique utterances, making several assumptions on which utterances will be treated as paraphrases of each other. Specifically, utterances with identical skill, intent, set of slots and having more than half of longer utterance words in common were treated as paraphrases and merged in groups, since they have identical in intent, entities, required actions from Alexa perspective. Having half words in common reduces the noise, since often skills and intents are rather generic and consist of mostly non-entity words, so the utterances might be very different. For example, sentences "how much does it cost to rent \textless CarSlot\textgreater" and "can i rent a \textless CarSlot\textgreater" have only one slot in common and don't carry exactly the same meaning in general case. Often the utterances are much more noisy, and if they have no slots, they are ignored. Eventually we take the permutations within each group to form our paraphrase corpus with 1M pairs.\\

The in-domain training data consists of Alexa built-in domains and ASK skill utterances. In this paper, we are using a subset of skills as a test set, while all other data is used for the training and validation sets. This dataset sentences tend to be very short (usually 2-7 words) and are not very diverse. Given that and  moderate size of the corpus, it might not be feasible to use it for training encoder-decoder network directly and get paraphrases of satisfying quality. Thus we are using out-of-domain parallel translation corpora for encoder training - MultiUN \ccite{chen2012multiun} and Europarl \ccite{koehn2005europarl}, with 7M training sentences aligned for English, French and Spanish. The vocabulary is quite different from Alexa data and it does not have a lot of names, entities, words like "uber", "airfare" or "pokemon". However, this data allows to build stronger encoder.

\section{Model Description} \label{model-desc}

\begin{figure*}[h!]
     \centering
     \includegraphics[width=0.7\textwidth]{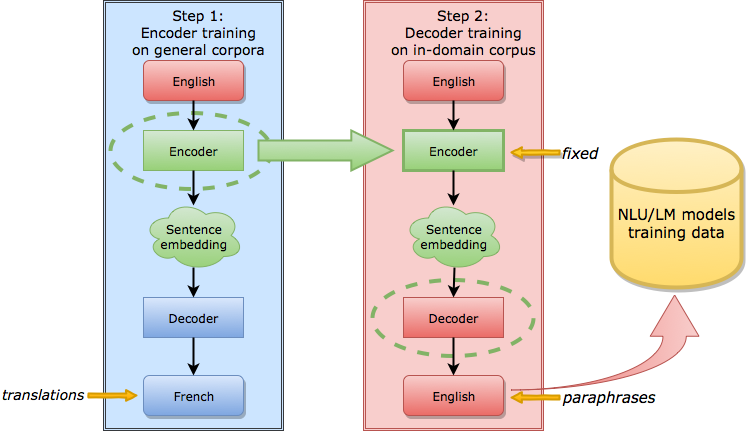}
     \caption{Paraphraser model training procedure. Steps 2 uses fixed encoder weights, trained on Step 1.}
     \label{fig:enfr}
\end{figure*}

We train our paraphraser model in two stages(see Figure \ref{fig:enfr}). First we are training encoder on large out-of-domain parallel corpora. Then we fix the encoder and retrain decoder on smaller in-domain data. Once the model is trained, multiple paraphrases can be generated for each given input example by using n-best decoding or sampling. The generated paraphrases are then combined with the original examples to build NLU and LM models for that skill.

It is important for model training that the out-of-domain corpora are multilingual parallel datasets. Neural networks in general can have enough parameters to learn mappings of variably high complexity. For natural language, it is possible that such mappings will carry either semantic or syntactic representation of a sentence, or combination of both. For instance, a model could learn one to one mappings between speech patterns or phrases. However, for paraphrase generation it is essential for the embedding to be semantic, i.e. to have enough representational capacity to ensure output diversity. If we train the encoder on different target languages, having different language structure, grammar and word order, we expect it to have much better chance of learning the sentence true semantic representation, than than the surface form.


\subsection{Copying Mechanism} \label{copy-mechanism}
\begin{figure}[h!]
     \centering
     \includegraphics[width=0.45\textwidth]{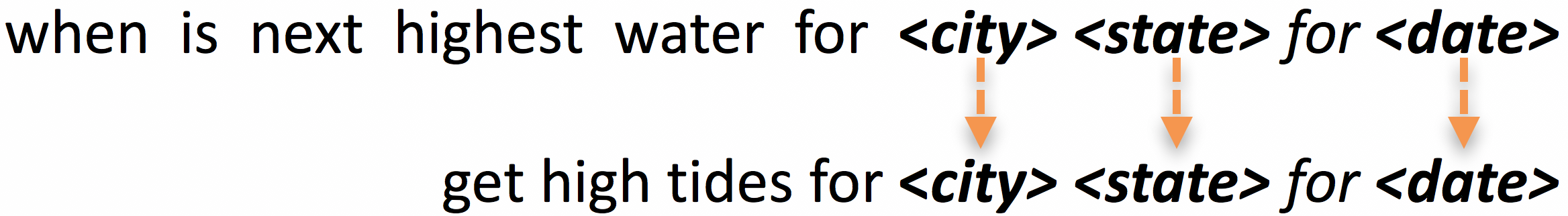}
     \caption{Example of desired slot copying behavior for utterance from "Tide pooler" skill. The same slots have to be present on the output to be replaced with the original slots contents afterwards.}
     \label{fig:slot-example}
\end{figure}

Figure \ref{fig:slot-example} shows example sentences from skill grammar. Note that the sentences contain slots (entity types), which are essential for LM and NLU models and must be preserved in the paraphrases. Also we do not want to paraphrase entities, such as movie titles, but to keep the original ones. Therefore, we have to come up with some sort of copying mechanism. We can not condition the encoder on the slots explicitly, since it is trained on larger out-of-domain corpora (see \nameref{data} section). So without introducing any changes to the model, we are implementing a copying mechanism through data pre-processing and post-processing. We are trying to replace all occurrences of slots with several levels of abstractions. The assumption is that a model, that sees these abstract tokens so often on both input and output, should learn to encode this information into embedding and propagate it to the output with high probability. There is certainly a trade-off of sacrificing the original words sequences with the context (i.e. “city of Seattle”) in favor of more abstract entities. This results in some degradation of the language model, however, it gives us the slots we necessarily need for ASK. During decoding we ignore paraphrases with different number or type of slots than in the input sentence to reduce the number of incorrect paraphrases.

\begin{figure}[h!]
     \centering
     \includegraphics[width=0.5\textwidth]{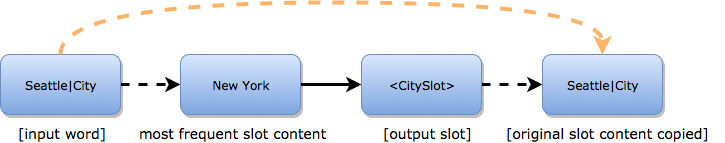}
     \caption{Copying mechanism example.  The original slot (Seattle \rule[-0.4ex]{0.2ex}{1em} City) replaces \textless CitySlot\textgreater on the generated sentence. Dashed arrows are pre/post-processing operations. Solid arrow is the actual model output.}
     \label{fig:copying}
\end{figure}

Figure \ref{fig:copying} demonstrates our approach. For in-domain data we replace the original input slot values (Seattle) with the most frequent word for that slot (New York), ignoring stop words. The target output slots contents (Seattle) are replaced with the actual slots (\textless CitySlot\textgreater - one of special tokens we add to output vocabulary). During decoding the generated output slot (\textless CitySlot\textgreater) is replaced with corresponding input slot (Seattle) if the slots match. Otherwise we reject the candidate paraphrase.

\subsection{Proposed Models}

\begin{figure}[h!]
     \centering
     \includegraphics[width=0.55\textwidth]{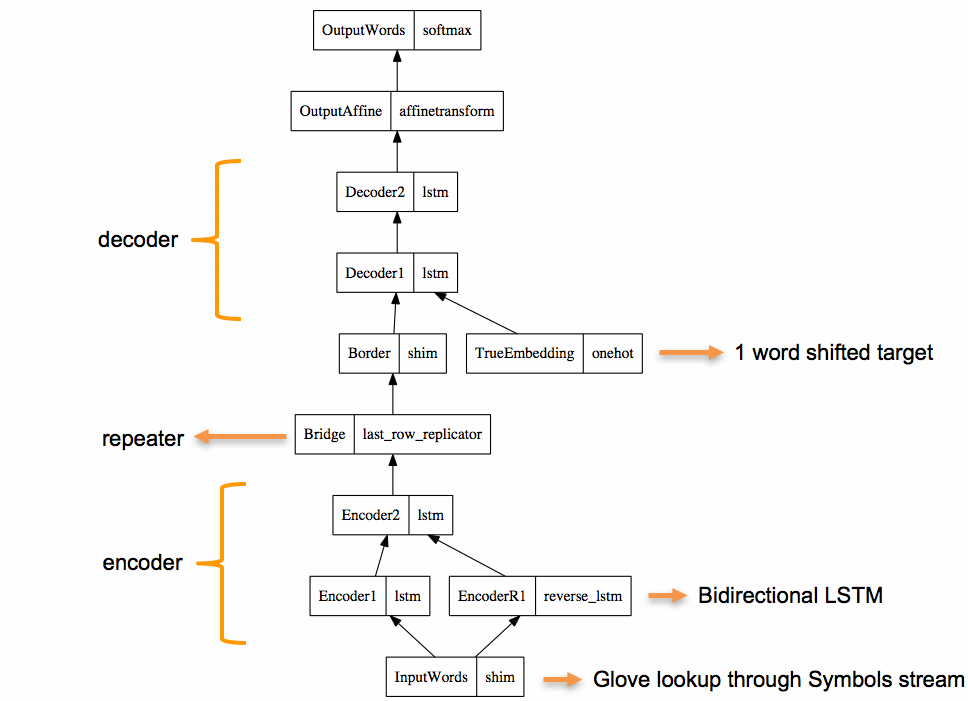}
     \caption{Model architecture in NeMo \ccite{strom2015scalable} - Amazon deep learning framework for distributed model training, easily scalable to many parallel GPU instances. Each block is a component labeled with its name and type.}
     \label{fig:model}
\end{figure}

The proposed model is shown in Figure \ref{fig:model}. It is following architecture by Cho \ccite{cho2014properties} with the only difference in using GloVe word-embeddings and bidirectional LSTM as the first layer.
It has 11M parameters. The input to the model is a sequence of 300-dimensional GloVe vector embedding,  looked up from a dictionary for each token.
Encoder consists of one bidirectional and one regular LSTM, last hidden state of which is the encoder embedding. At every time step the two-layer decoder LSTM takes that embedding as well as the previously generated word one hot encoding and feeds it into fully-connected layer, followed by a softmax.
 
First, we train standard MT model on English to French parallel corpus. Then we continue training on in-domain English paraphrase corpus, following three different training schemes:
\begin{center}
\begin{itemize}
\item No slot copying - don't apply slot copying on the in-domain data, i.e. keep the slot content on both input and output. Fix encoder weights and retrain decoder on this data.
\item Fix encoder - apply slot copying on the in-domain data. Fix encoder weights and retrain decoder on this data. Do output post-processing to replace slots with the original slots contents.
\item Fine-tune encoder - apply slot copying on the in-domain data. Retrain both encoder and decoder on this data. Do output post-processing to replace slots with the original slots contents.
\end{itemize}
\end{center}


\section{Results} \label{results}

\subsection{Evaluation Metrics}

We evaluate our models on the effect the generated paraphrases bring to the ASK models on the unseen skills. We are sampling the skill data from its grammar and apply corresponding copying mechanism. We might also generate the input from grammar directly instead.
Our main evaluation criteria is to compare the accuracy of a skill baseline models with the skill models trained on the paraphrase augmented data. The baselines are the existing NLU models - maximum entropy model for intent classification (IC) and linear chain conditional random field for named entity recognition (NER). Both are trained solely on the data sampled from skill grammar. Slot error rate (SER) is calculated as word error rate, but only for slots. Semantic error rate (SEMER) is the ratio of number of errors to number of reference slots:
$SEMER = \frac{S+I+D}{S+D+C}$,
where $S$ and $C$ are numbers of substituted and correct slots and intents, $I$ and $D$ are numbers of slot insertions and deletions correspondingly.
                                               
We also introduce diversity metrics, including word and sentence level coverage of live data by the original and augmented training data. Word coverage is the ratio of total number of unique words. Sentence coverage is the average ratio of bigrams per sentence.

\subsection{Quantitative Results}
\begin{figure}[h!]
     \centering
     \includegraphics[width=0.45\textwidth]{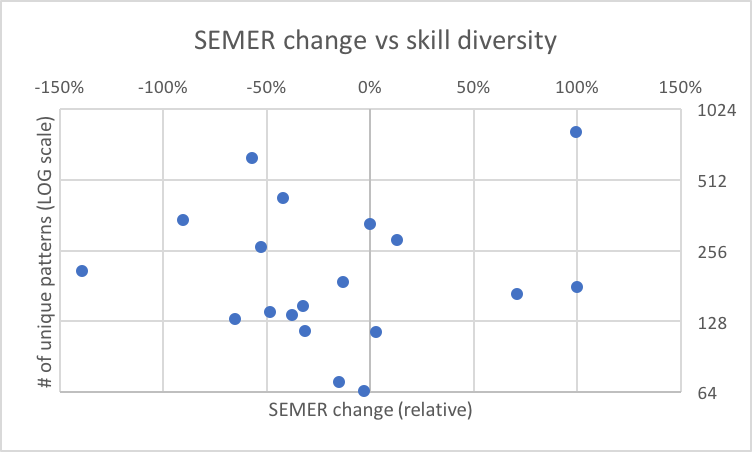}
     \caption{Skill SEMER change by skill diversity for training data augmented with fixed encoder model. Skill diversity is calculated as the number of unique patterns in annotated live data.}
     \label{fig:semer_usage}
\end{figure}

All tested skills have four or more times improvement on the bigram coverage score, compared to the corresponding skills baselines. Data augmentation has better effect on the IC than SER. Figure \ref{fig:semer_usage} shows the skills relative SEMER change versus the skill usage. Data augmentation was done with the fixed encoder model trained on out-of-domain corpus. Table \ref{tbl:stubhub} shows all the evaluation metrics we are using over different models for a single skill - Stubhub.

\begin{table*}
\centering
\caption{Paraphrases evaluation on Stubhub skill. Baseline has the results for ASK models trained on the sampled grammar only. Other models are augmenting the samples with paraphrases. Test dataset (annotated live data) consists of 635 utterances. Input/output size is the number of original and generated utterances.}
\label{tbl:stubhub}
\begin{tabular}{|c|c|c|c|c|c|}
\hline
\multirow{2}{*}{\textbf{Model}} & \multirow{2}{*}{\textbf{Bigram coverage}} & \multirow{2}{*}{\textbf{\begin{tabular}[c]{@{}c@{}}Input/output\\ size\end{tabular}}} & \multicolumn{3}{c|}{\textbf{Relative error change over baseline}} \\ \cline{4-6} 
                                  &                                             &                                                             & \textbf{ICER}      & \textbf{SER}     & \textbf{SEMER}    \\ \hline
baseline                          & 0.036                                       & 490/0                                                       & 0\%                  & 0\%                & 0\%                 \\ \hline
no slot copying                        & 0.271                                       & 490/1764                                                    & -14.3\%              & +44.1\%            & +9.6\%              \\ \hline
fixed encoder                     & 0.124                                       & 63/559                                                      & \textbf{-26.6\%}   & \textbf{+2.5\%}  & \textbf{-13.5\%}  \\ \hline
fine-tuned encoder                   & 0.101                                       & 64/350                                                      & -15.9\%              & +17.9\%            & -2.2\%              \\ \hline
\end{tabular}
\end{table*}

\subsection{Qualitative Analysis}
Unlike French translations, English paraphrases don't have any unknown words, even if the copying mechanism is not used. This must be due to the fact that Alexa users use less rich vocabulary that the European parliament. Small vocabulary and  much shorter average length of In-domain sentences result in smaller word error rate and better translation quality, compared to the out-of-domain corpora.
The model does not differentiate between numbers and entities, i.e. cities, and uses them interchangeably. This is because the entities are in close proximity in the embedding space. And since it only increases the output diversity, the effect is very much desired. Though it is not an actual paraphrase, since the quantity or location is different, it would not corrupt the training data, since the intent and slots are the same.
The paraphrase model trained without copying mechanism produces better grammar sentences and sometimes manages to propagate correct entities to the output.

\section{Discussion and future work} \label{discussion}

As we have shown, augmenting the training data with paraphrases using our model helps to significantly improve Alexa IC and NER. The paraphrases quality and grammar, however, suffers due to the trade-offs we have done. For the existing model as it currently is, it is possible that careful optimization and paraphrase selection will yield high quality paraphrases, which will be enough for improving IC accuracy across majority of the skills, especially since it is pretty robust to noise, if provided with proper slots. However, NER and potentially other models, which can use the paraphraser would need less noisy and more reliable model.

Slots copying mechanism through post-processing is by no means optimal approach to entity tagging. We have chosen it as the simplest solution in the given time constraints. It often does not keep the semantics and the context of the sentence, as the input entities are abstracted out, resulting in more noise. Models using the copying mechanism tend to have worse grammar than the models not using it. 
The latter models work very reasonably for general data, however, we can not use it for ASK data augmentation, which necessarily needs proper slots in the paraphrases. 

In the future, we plan to explore other approaches to slot copying problem without clustering entities into abstractions.
One option is tagging the output entities as post-processing step, but that would require the model to propagate those entities there. 
We can not condition the encoder on the slots to give the model additional signal, since encoder is trained on out-of-domain data. 
A workaround would be to annotate the out of domain corpus with generic entities and then mapping it to in-domain slots.
More interesting solution would be conditioning the decoder on the input slot sequence instead. It is completely decoupled from encoder training and we would make use of the slots we are provided on the input.




Another problem of the proposed model is decoupling the training data into disjoint French and English corpora. We would want to get more representative parallel data, which is closer to our data distribution. This can be done by tagging existing French corpus or translating the in-domain data to French, preferably with slots. We might also use OpenSubtitles \ccite{lison2016opensubtitles2016} corpus, which might correlate much better with Alexa data, since it mostly consist of movie dialogues.


Other things we are considering is changing the decoding procedure to sample the hypotheses using posterior probabilities in order to increase output diversity. We might also use a dedicated paraphrase selection model, i.e. for reranking the paraphrases. There is definitely some room left for optimization and adding standard NMT enhancements as attention mechanism \ccite{bahdanau2014neural} or deeper stacked LSTMs with residual connections \ccite{prakash2016neural}.

\section{Conclusion} \label{conclusion}

In this work, we propose a novel approach to paraphrasing, and a training protocol that allows us to exploit large amount of out-of-domain data. We also propose a method for paraphrasing sentences with slots. We demonstrate the effectiveness of our model NER and IC tasks, showing substantial improvement over the baseline. This allows us to reduce the amount of needed manual annotations, and make it easier for developers to create high-quality skill grammars.


\bibliographystyle{plainnat}
\newpage
\bibliography{paraphraser}

\end{document}